\documentclass{article}

\usepackage{microtype}
\usepackage{graphicx}
\usepackage{subfigure}
\usepackage{booktabs} 

\usepackage{amsmath,amssymb,amsfonts}
\usepackage{amssymb}
\usepackage{algorithmic}
\usepackage{textcomp}
\usepackage{multirow}
\usepackage{enumitem}
\usepackage{textcomp}
\usepackage{makecell}
\usepackage{multirow}
\usepackage{tabularray}
\usepackage{array, makecell}

\usepackage{float}
\usepackage{fancyhdr}
\usepackage{colortbl}
\usepackage{arydshln}
\usepackage{graphicx}
\usepackage{textcomp}
\usepackage{soul}
\usepackage{hyperref}
\usepackage{tcolorbox}
\usepackage{wrapfig}
\usepackage{amssymb}
\usepackage{pifont}
\usepackage{graphicx}
\usepackage{textcomp}
\usepackage{xcolor}
\usepackage{makecell}
\usepackage{multirow}
\usepackage{tikz}
\usepackage{float}
\usepackage{graphicx}
\usepackage{booktabs}
\usepackage{cleveref}
\crefformat{section}{\S#2#1#3} 
\crefformat{subsection}{\S#2#1#3}
\crefformat{subsubsection}{\S#2#1#3}

\usepackage{hyperref}
\usepackage{microtype}
\usepackage{subfigure}
\usepackage{booktabs} 

\usepackage{pifont}
\usepackage{cite}
\usepackage{newtxtext,newtxmath}
\usepackage{textcomp}
\usepackage{comment}
\usepackage{fancyhdr}
\usepackage{lipsum}
\usepackage{comment}
\usepackage{outlines}
\usepackage{graphicx}
\usepackage{float}
\usepackage{tablefootnote}
\usepackage[export]{adjustbox}

\usepackage{multirow}
\usepackage{pifont}

\usepackage{subcaption}

\usepackage{svg}
\usepackage{hyperref}

\usepackage{forest}
\usepackage{subcaption}

\newcommand*\circled[1]{\tikz[baseline=(char.base)]{\node[shape=circle,fill,inner sep=0pt] (char) {\textcolor{white}{#1}};}}

\definecolor{revA}{HTML}{000000}
\definecolor{revB}{HTML}{000000}
\definecolor{revC}{HTML}{000000}
\definecolor{revD}{HTML}{000000}
\definecolor{revE}{HTML}{000000}

\definecolor{colorRevA}{rgb}{0.85, 0.90, 0.98}
\definecolor{colorRevB}{rgb}{0.97, 0.80, 0.796}
\definecolor{colorRevC}{rgb}{0.87, 0.83, 0.90}
\definecolor{colorRevD}{rgb}{0.83, 0.91, 0.83}
\definecolor{colorRevE}{rgb}{1, 0.90, 0.80}
\definecolor{blond}{rgb}{0.98, 0.94, 0.75}

\usepackage[accepted]{mlsys2025}

\mlsystitlerunning{Scaling Multi-Node Mixture-of-Experts Inference Using Expert Activation Patterns}

\begin{document}

\usetikzlibrary{shadows, arrows.meta}

\tikzset{parent/.style={align=center,text width=2cm,fill=green!20,rounded corners=2pt},
    child/.style={align=center,text width=2.8cm,fill=green!50,rounded corners=6pt},
    grandchild/.style={fill=pink!50,text width=2.3cm}
}

\newcommand{\SubItem}[1]{
    {\setlength\itemindent{15pt} \item[-] #1}
}
\newcommand{\abc}{{\sf abc}\xspace}
\newcommand{\dse}{{\sf dse}\xspace}
\newcommand{\defns}{{\sf def}}
\newcommand{\tool}{{ELAP }}
\newcommand{\toolns}{{ELAP}}

\newcommand{\AB}[1]{{\color{magenta}\bfseries [Abhi:: #1]}}
\newcommand{\RR}[1]{{\color{purple}\bfseries [Ritik:: #1]}} 
\newcommand{\GJ}[1]{{\color{blue}\bfseries [Geonhwa:: #1]}} 
\newcommand{\SudS}[1]{{\color{orange}\bfseries [Sudarshan:: #1]}}
\newcommand{\SK}[1]{{\color{purple}\bfseries [Souvik:: #1]}} 
\newcommand{\TK}[1]{{\color{red}\bfseries [Tushar:: #1]}}
\newcommand{\TODO}[1]{\textcolor{red}{TODO:: #1}}
\newcommand{\rev}[1]{\textcolor{blue}{#1}}
\newcommand{\fixme}[1]{\textcolor{red}{#1 \textit{- FIXME}}}

\newcommand{\niparagraph}[1]{\vspace{1pt}\noindent\textbf{#1}}
\newcommand{\checkbox}{\textbf{\textcolor{green}{\ding{51}}}}
\newcommand{\xbox}{\textbf{\textcolor{red}{\ding{55}}}}

\newcommand*\hcircled[1]{\tikz[baseline=(char.base)]{\node[shape=circle,draw,inner sep=2pt] (char) {#1};}}


\def\sectionautorefname{Section}
\def\subsectionautorefname{Section} \def\subsubsectionautorefname{Section }

\def\figureautorefname{Fig.}

\definecolor{turqoise}{HTML}{4BCAD2}
\definecolor{orangered}{HTML}{CF1040}
\definecolor{redgreen}{HTML}{66AA00}
\definecolor{lightgray}{gray}{0.9}

\twocolumn[
\mlsystitle{Scaling Multi-Node Mixture-of-Experts Inference \\ Using Expert Activation Patterns}




\begin{mlsysauthorlist}

\mlsysauthor{Abhimanyu Bambhaniya}{gt}
\mlsysauthor{Geonhwa Jeong}{m}
\mlsysauthor{Jason Park}{m}
\mlsysauthor{Jiecao Yu}{m}
\mlsysauthor{Jaewon Lee}{m}
\mlsysauthor{Pengchao Wang}{m}
\mlsysauthor{Changkyu Kim}{m}
\mlsysauthor{Chunqiang Tang}{m}
\mlsysauthor{Tushar Krishna}{gt}

\mlsysaffiliation{gt}{Georgia Institute of Technology, USA}
\mlsysaffiliation{m}{Meta Platforms, Inc., USA}

\mlsyscorrespondingauthor{Abhimanyu Bambhaniya}{abambhaniya3@gatech.edu}
\end{mlsysauthorlist}



\begin{abstract}
Most recent state-of-the-art (SOTA) large language models (LLMs) use Mixture-of-Experts (MoE) architectures to scale model capacity without proportional per-token compute, enabling higher-quality outputs at manageable serving costs. 
However, MoE inference at scale is fundamentally bottlenecked by expert load imbalance and inefficient token routing, especially in multi-node deployments where tokens are not guaranteed to be routed to local experts, resulting in significant inter-node all-to-all communication overhead.









To systematically characterize these challenges, we profile SOTA open-source MoE models, including Llama 4 Maverick, DeepSeek V3-671B, and Qwen3-230B-A22B, on various datasets and collected over 100k real expert activation traces. Upon studying the expert activation patterns, we uncover various persistent properties across all the frontier MoE models: variable expert load imbalance, domain-specific expert activation where expert popularity shifts across task families (code, math, chat, general), and a strong correlation between prefill and decode expert activations.

Motivated by these findings,  we propose workload-aware micro-batch grouping and an expert placement strategy to maximize token locality to the destination expert, thereby reducing inter-node communication.
Across models and datasets, these optimizations help reduce all-to-all communication data up to \textbf{20\%}, resulting in lower MoE decode latency and better accelerator utilization.

%

\end{abstract}
]



\printAffiliationsAndNotice{}  

\section{Introduction}
The emergence of state-of-the-art Mixture-of-Experts (MoE) models, including DeepSeek-V3~\citep{deepseekai2025deepseekv3technicalreport}, Qwen3-MoE~\citep{qwen3technicalreport}, and Llama4~\citep{Llama4}, has fundamentally transformed the landscape of large language model inference. These models achieve remarkable performance with significantly reduced inference FLOPs by activating only a subset of experts per token~\citep{fedus2021switch}. However, the high sparsity that makes MoE models efficient at the algorithmic level introduces substantial challenges for multi-node distributed inference systems.

To meet the computational demands of these humongous MoE models, distributed inference systems need to leverage large expert parallelism (EP) combined with prefill-decode disaggregation across multiple nodes~\citep{deepseekai2025deepseekv3technicalreport}. This approach appears straightforward in principle: distribute experts across a cluster of nodes and route tokens to their designated experts. However, empirical observations reveal a counter-intuitive reality; naively scaling expert parallelism to larger node counts often fails to deliver expected performance improvements, particularly during the decode phase where latency sensitivity is critical.

\begin{figure}[t]
\centerline{\includegraphics[width=\linewidth]{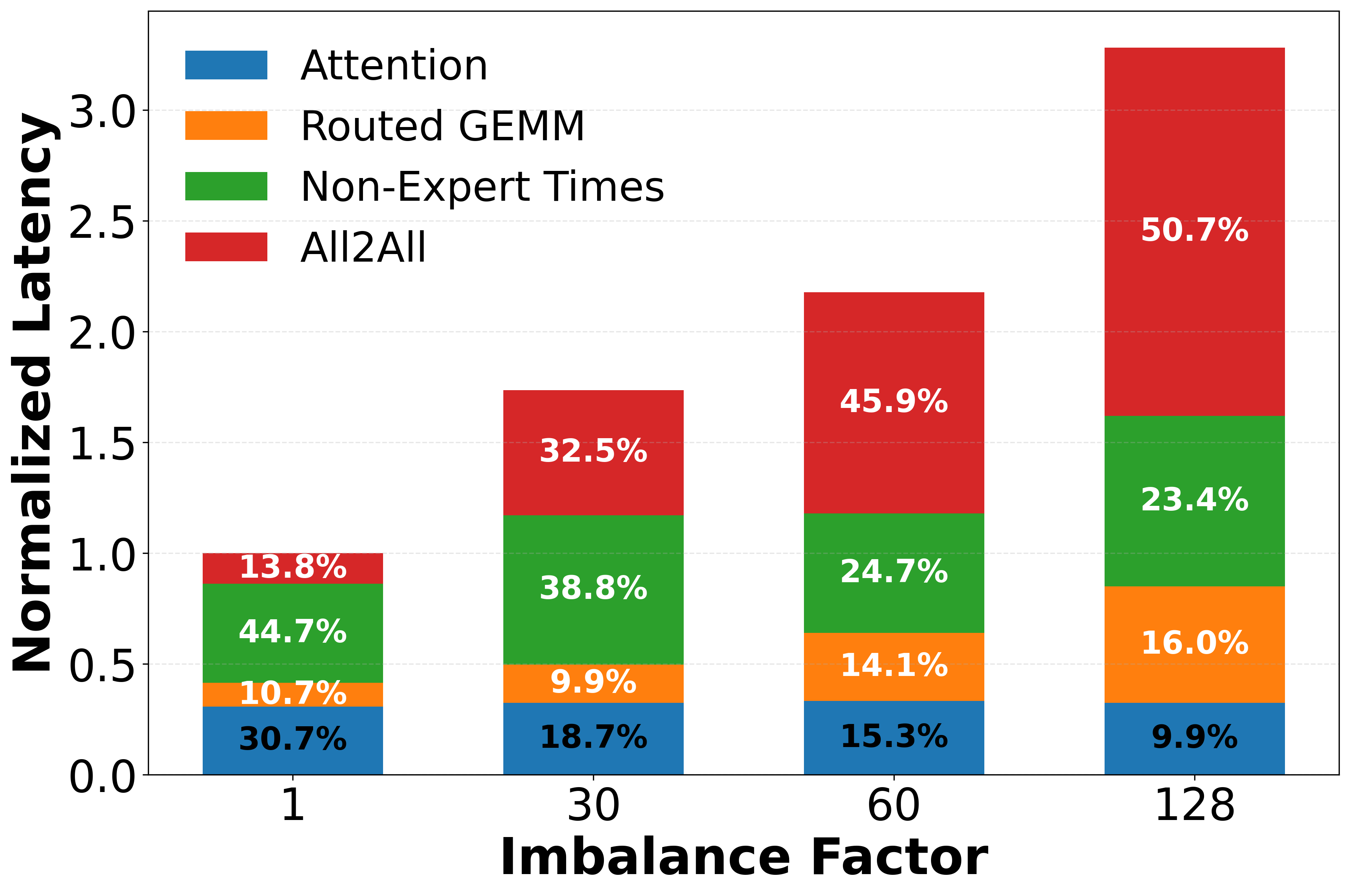}}
\caption{
Multi-Node Llama-Maverick Decode Latency Breakdown. As imbalance between EP ranks increase, the all-to-all component raises significantly.
The text in each box in a bar is \% of layer runtime compared the total runtime for each bar.
}
\label{fig:moe_latency_breakdown}
\end{figure}

The fundamental challenge lies in the \textbf{mismatch between expert routing patterns and physical hardware topology}. Unlike dense models where computation is predictably distributed thanks to their deterministic computation patterns, MoE models exhibit highly dynamic computation patterns due to the input-dependent expert activation patterns. This is critical during multi-node MoE inference as tokens frequently need to be routed to expert located on different nodes, triggering expensive inter-node all-to-all communication. As shown in \autoref{fig:moe_latency_breakdown}, this all-to-all communication overhead becomes the dominant bottleneck, often overwhelming compute/memory benefits from large parallelism. In addition, the discrete nature of expert routing decisions creates a combinatorial optimization problem: even small changes in expert placement can dramatically affect communication patterns.

%

Existing approaches fall short of addressing this non-trivial challenge comprehensively. Linear placement strategies~\citep{vLLM_expert_placement_2025} ignore activation patterns entirely and just follow the order of expert id in the model when placing the experts on GPUs. 
DeepSeek introduced EPLB approaches~\citep{deepseek_EPLB_2025_github} using historical load statistics to balance expert loads across GPUs. However, it fails to capture fine-grained token-expert routing patterns. Neither approach considers the critical insight that requests with similar expert activation patterns should be co-located and processed together to maximize intra-node computation while minimizing inter-node communication.

\begin{figure}[t]
  \centering
    \subfigure[Deepseek-V3]{\includegraphics[width=0.48\columnwidth]{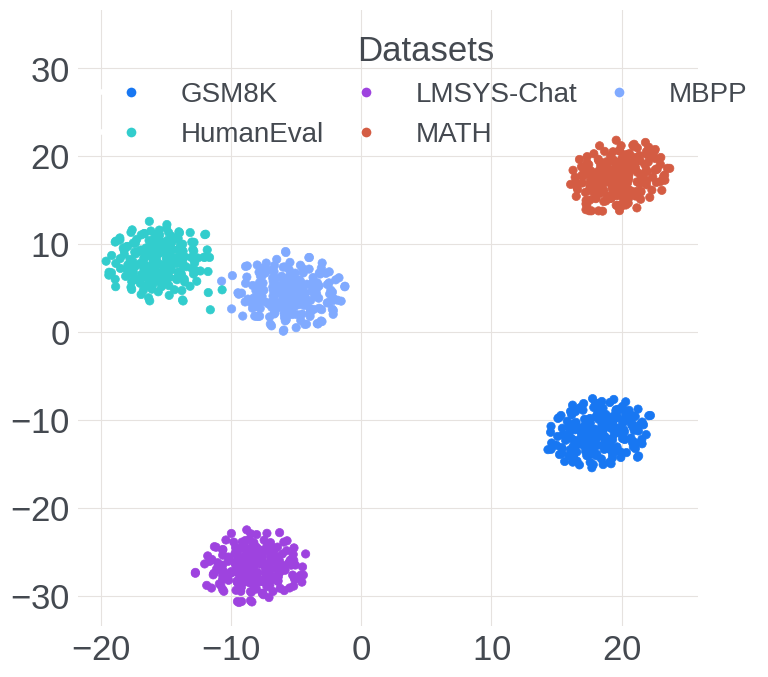}} 
    \subfigure[Qwen3-235B-A22B]{\includegraphics[width=0.48\columnwidth]{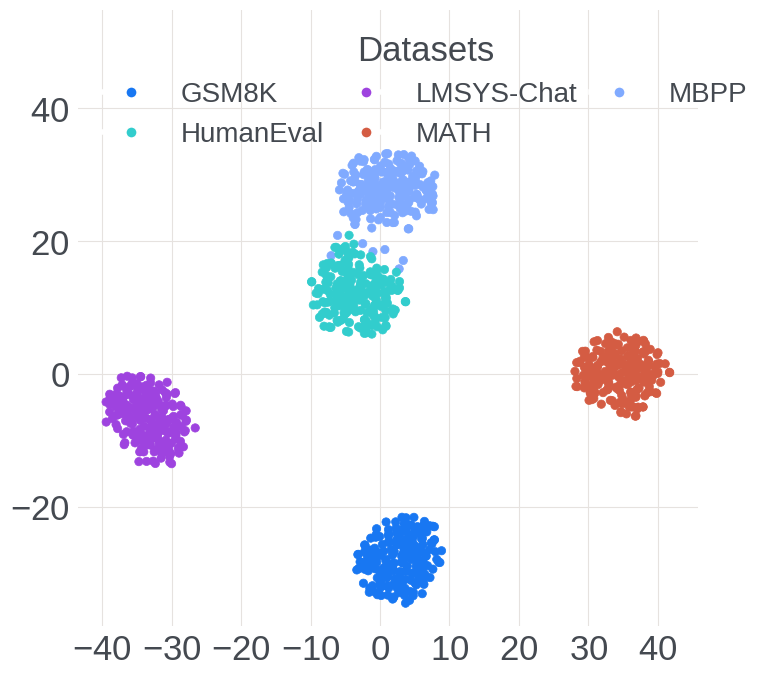}} 
  \caption{Dataset wise decode expert activation visualization.}
  \label{fig:dataset_grouping}
\end{figure}

To address these fundamental limitations, we introduce a data-driven approach that leverages comprehensive analysis of expert activation patterns to optimize both request batching and expert placement in multi-node MoE inference serving systems. Our method is grounded in empirical observations from over 100,000 inference requests across three state-of-the-art MoE models, revealing that expert activation patterns exhibit predictable clustering (\autoref{fig:dataset_grouping}) properties that can be exploited for system optimization.

Based on our analysis, we propose two key innovations. First, we perform workload-aware micro-batch grouping by clustering requests after prefill based on their expert activation patterns during prefill, ensuring that co-batched requests activate similar sets of experts. Second, we develop an expert placement algorithm that strategically places experts across device groups for decode to maximize the fraction of tokens that can be processed within single nodes, thereby minimizing expensive inter-node all-to-all communication.

Our contributions are threefold: 
\begin{enumerate}
    \item We present the first comprehensive characterization of expert activation patterns across multiple state-of-the-art MoE models and diverse datasets, revealing critical insights about load imbalance, activation correlation, and clustering properties. 
    \item We design a principled approach that combines expert-activation-aware request clustering with strategic expert placement to minimize inter-node communication overhead. 
    \item We demonstrate that our clustering and placement approach reduces all-to-all communication volume up to 20\% and reduce MoE layer latency by 6\% compared to existing placement strategies.
\end{enumerate}

This work addresses a critical gap in scaling MoE inference to large multi-node clusters, providing both theoretical insights and practical solutions for the next generation of sparse language models.


\section{Background}
\label{sec:background}

\subsection{Mixture-of-Experts}
Mixture-of-Experts (MoE) models~\citep{GPT4,fedus2021switch, mixtral22b, grok, kimiteam2025kimik2openagentic} have emerged as a dominant architecture for scaling Transformer-based language models. Unlike dense models, MoEs have parallel feed-forward networks (experts), with a routing mechanism that dynamically selects a subset of experts for each input token. This design enables significant increases in model capacity with only modest increases in inference cost, as only a fraction of the model is activated per token.
Recent state-of-the-art MoE models, such as  DeepSeek-V3~\citep{deepseekai2025deepseekv3technicalreport}, Qwen3-MoE~\citep{qwen3_blog}, and Llama Maverick~\citep{Llama4} have demonstrated that MoE architectures can match or exceed the performance of dense models at a fraction of the inference FLOPs.
\subsubsection{Attention}
MoE models typically retain the standard self-attention mechanism of Transformers. The MoE block is usually inserted after the attention layer, replacing or augmenting the standard feed-forward network (FFN).
\subsubsection{Routing}
The core of MoE inference is the routing mechanism, which determines which experts process each token. The most common approach is expert choice routing top-$k$ gating~\citep{deepseekai2025deepseekv3technicalreport,mixtral22b}, where each token is routed to the $k$ experts with the highest gating scores. 
A key challenge with routing is load imbalance: the distribution of tokens across experts can be highly skewed, leading to stragglers experts as the highest loaded experts can become bottleneck keeping significant hardware resources underutilized. While auxiliary losses during training can encourage balanced routing~\citep{lepikhin2021gshard,fedus2021switch,zoph2022stmoedesigningstabletransferable}, inference-time imbalance remains a practical concern, especially at datacenter scale deployments.
\subsubsection{Routed Experts}
\emph{Routed experts} only process tokens selected by the gating mechanism. As a result, the number of tokens per expert is variable and can be unpredictable, complicating load balancing and parallelization.

\subsection{Parallelism for MoE Models}
Large model inference of large MoE models requires distributing computation across multiple devices and nodes. Three primary types of parallelism are employed for MoE inference:
\subsubsection{Data Parallelism}
Data parallelism replicates the model parameters across devices, each processing a different batch of input tokens. 

\subsubsection{Tensor Parallelism}
Tensor parallelism~\citep{shoeybi2019megatron} splits the computation of large matrix multiplications (e.g., within the FFN or attention layers) across devices, typically along the hidden dimension. This is essential for fitting large models into device memory and for scaling up single-model inference.
\subsubsection{Expert Parallelism}
Expert parallelism~\citep{lepikhin2021gshard} distributes the set of experts across devices. Each device is responsible for a subset of experts, and tokens are routed to the appropriate device based on the gating decision. This is the most prominent form of parallelism for MoE models, but introduces all-to-all communication overhead due to token-expert assignments.

\subsection{Hybrid Parallelism}
\label{sec:hybrid-parallelism}
DeepSpeed-MoE~\citep{rajbhandari2022deepspeedmoeadvancingmixtureofexpertsinference} introduced hybrid parallelism strategy, combining distinct forms of parallelism to expert and non-expert parameters to scale the number of nodes while reducing under-utilization


\paragraph{Non-Expert Parameters (Attention, Shared FFN)}
Non-expert parameters, such as those in attention layers and shared feed-forward networks, are required by all tokens. For these components, we use \textbf{tensor parallelism} (TP) within a group.
For LLM inference decode stage, it is important to have large batch sizes to avoid under-utilization. 
For scaling the batch size, non-expert parameters are replicated multiple times via \textbf{data parallelism} (DP). Each DP group maintains a replica of the non-expert parameters and processes different batches of input data, eliminating inter-group communication for non-expert layers. 

\paragraph{Expert Parallelism for Routed Experts}
Routed Experts in each MoE layer, are partitioned using \textbf{expert parallelism} (EP). Each device is assigned a subset of experts, and all tokens routed to a given expert are processed together on the corresponding device. This grouping minimizes the critical data path per device and enables high throughput.
To further scale beyond the number of available experts, we can utilize \textbf{tensor parallelism} (TP), wherein the parameters of each expert are themselves partitioned (horizontally or vertically) across multiple GPUs.

Combining DP+TP for non-expert parameters and EP+TP$_{exp}$ for expert parameters we can achieves scalable, low-latency, and high-throughput inference system for multi-trillion parameter MoE models. We refer these scheme as DPaTPb$\rightarrow$EPcTPd throughout the paper, ex: DP8TP8$\rightarrow$EP32TP2, mean 64 GPU used with 8 replicas of non-expert parameters shared in 8 GPUs and expert parameters are distributed into 32 EP rank with each rank splits the experts into 2 nodes.

Recent works have proposed various strategies for expert placement~\citep{efficientMoE_inference_fair}, token permutation~\citep{gale2022megablocksefficientsparsetraining,efficientMoE_inference_fair}, and communication~\citep{jin2025megascalemoelargescalecommunicationefficienttraining,deepseekai2025deepseekv3technicalreport}. However, achieving near-linear scaling for MoE inference across multi-node clusters remains an open problem.

\section{Characterization of Expert Activation}
\label{sec:characterization}
In this section, we analyze expert activation patterns in large-scale MoE models across diverse datasets and inference stages. We construct expert activation dataset, and present data-driven insights on load imbalance, activation correlation, prefill–decode predictability, and clustering. These results highlight core bottlenecks, optimization opportunities, and security risks in multi-node MoE inference.
\begin{table}[!bthp]
\caption{Architectural Parameters of SOTA MoE models.}
\label{tab:model_params}
\resizebox{\linewidth}{!}{\begin{tabular}{|c|c|c|c|}
\hline
\textbf{Feature} & \textbf{\begin{tabular}[c]{@{}c@{}}DeepSeek\\ V3\end{tabular}} & \textbf{\begin{tabular}[c]{@{}c@{}}Llama-4\\ Maverick\end{tabular}} & \textbf{\begin{tabular}[c]{@{}c@{}}Qwen 3\\ 235B-A22B\end{tabular}} \\ \hline
\textbf{Total Size} & 671B & 400B & 235B \\ \hline
\textbf{Active Size} & 37B & 17B & 22B \\ \hline
\textbf{\begin{tabular}[c]{@{}c@{}}\# Layers \\ (MoE Layers)\end{tabular}} & 61 (58) & 48 (24) & 94 \\ \hline
\textbf{Experts per Layer} & 256 & 128 & 128 \\ \hline
\textbf{Router (Expert/Token)} & 8 & 1 & 8 \\ \hline
\textbf{Shared Experts} & Yes & Yes & No \\ \hline
\end{tabular}
}
\end{table}
\subsection{Expert Activation Dataset Generation}

We collect MoE expert activation traces from over 100,000 requests spanning three large-scale models: Llama-4-Maverick~\citep{Llama4}, DeepSeek-V3~\citep{deepseekai2025deepseekv3technicalreport}, Qwen3-235B-A22B~\citep{qwen3technicalreport}, whose architectural parameters are summarized in \autoref{tab:model_params}. All models are evaluated as released, with FP8 quantization applied for inference. We implemented activation logging on top of vLLM v0.9~\citep{Vllm_Project_2025} and instrumenting the fusedMoE router to record, for each token in every request, the dataset, request index, decoding stage, input sequence length, generated token count, layer index, and a mapping of expert indices to token counts. Logging is performed at batch size 1 to ensure per-token granularity.
To cover broad set of datasets, we used GSM8K~\citep{cobbe2021gsm8k},
MATH~\citep{hendrycksmath2021},
MBPP~\citep{austin2021program},
HumanEval~\citep{chen2021evaluating},
LMSYS-Chat~\citep{zheng2023lmsyschat1m},
XSUM~\citep{Narayan2018DontGM}, and
TL;DR ~\citep{vonwerra2022trl}, with datasets approximately equally represented.

Data collection was conducted over 300 hours using 8 H100 GPUs.
No prompt inputs or outputs are logged, fully mitigating privacy concerns. 

\subsection{Terminology}
\subsubsection{Expert Load} 
We define Expert load as the number of tokens per expert divided by the number of perfect-load-balanced-tokens per expert.
If $expert\_load$ = 1, the expert takes the expected amount of tokens.
This model has 16 experts with top-1, so it can go up to 16 (i.e. when all tokens are routed to a single expert).

\subsubsection{Imbalance Factor}
Imbalance Factor (IF) of a layer = max(Expert load) per layer
If the imbalance factor equals to 1, the most “loaded” expert of the layer takes the expected amount of tokens.
Performance is bounded by the most loaded experts, and we will use this for performance projection.

\subsection{Expert Activation Analysis and Insights}
\subsubsection{Dataset-wise Load Imbalance}
\begin{figure}[!bthp]
    \centering
    \includegraphics[width=1\linewidth]{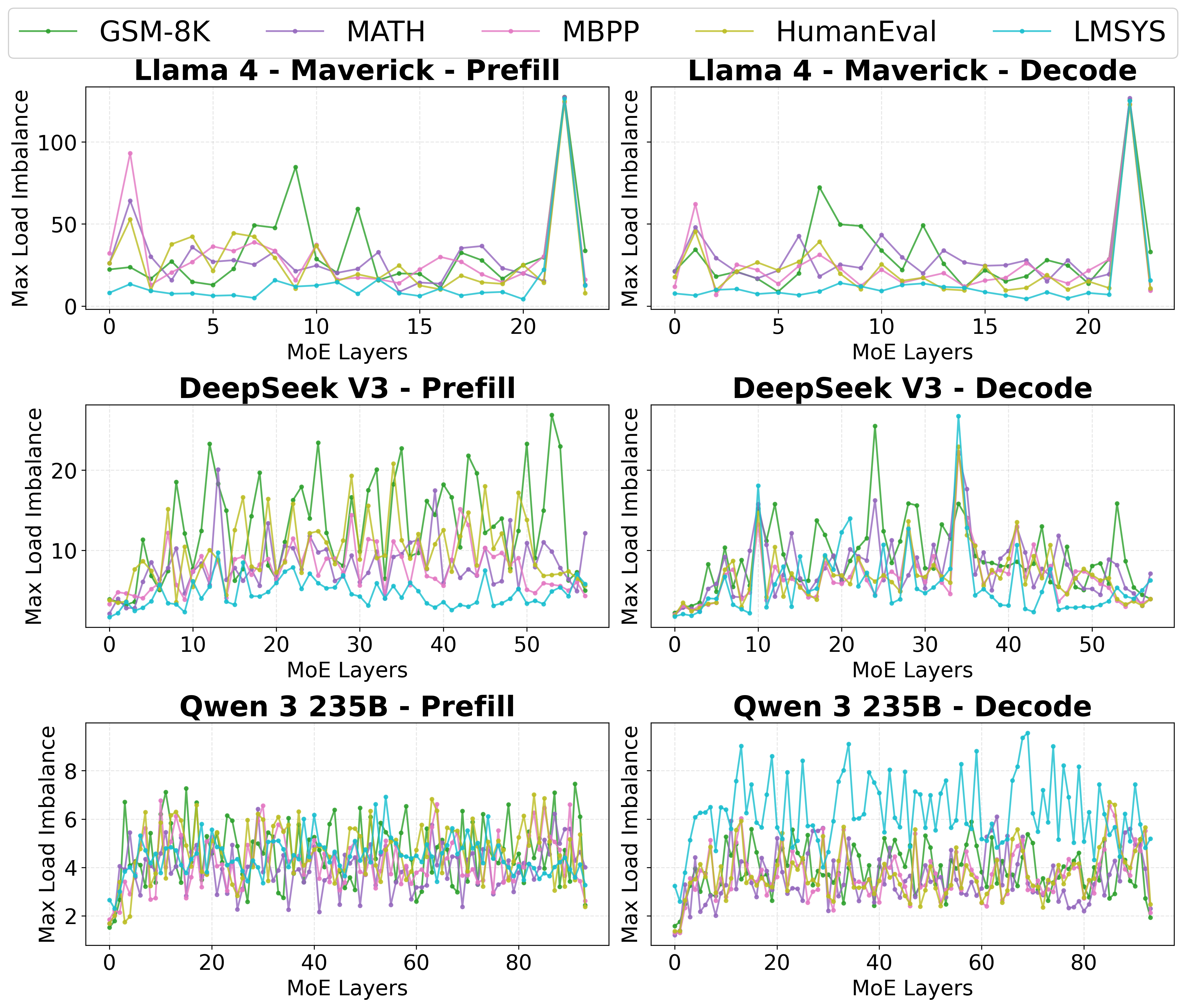}
    \caption{Maximum Load Imbalance  across MoE layers for different models  and datasets during prefill and decode phases.}
    \label{fig:load_imbalance_plots}
\end{figure}

Load imbalance across datasets and models are a critical bottleneck for efficient MoE inference, especially in multi-node settings. As seen in \autoref{fig:load_imbalance_plots}, the imbalance factor varies substantially by both dataset and model architecture.

Empirically, certain datasets induce pronounced imbalance in specific models and layers. For example, GSM8K consistently exhibits extreme imbalance in Llama 4-Maverick, with maximum expert loads exceeding 80$\times$ the balanced ideal.\footnote{Most of the tokens are routed to a single expert in Layer 45 (MoE layer 22), resulting in very high imbalance. This behavior has also been observed by other LLM Practitioners ~\citep{Meta_Ai_2025_bug}.
}
Such skewed token routing causes a small subset of experts to become bottlenecks during routed FFN computation. In contrast, unspecialized datasets like LMSYS-Chat show much lower imbalance—typically 10–20$\times$ in Maverick, and below 10$\times$ in DeepSeek V3 and Qwen3.
The root cause is dataset-dependent activation: homogeneous or repetitive tasks (e.g., GSM8K) activate a narrow set of experts, while diverse datasets distribute load more evenly. Notably, Qwen 3 235B maintains low imbalance across all datasets, with maximum expert loads rarely exceeding 6–8$\times$, indicating improved routing or more uniform expert specialization.
High imbalance, especially during prefill, leads to stragglers and hardware underutilization, bounding throughput by the slowest expert. This underscores the need for dataset-aware routing or adaptive load balancing for scalable MoE inference.
\begin{tcolorbox}[colback=green!10!white, colframe=green!80!black, title=Insight: Load Imbalance Drives Bottlenecks]
\textbf{Load imbalance is highly dataset- and model-dependent.} Specialized prompts (e.g., math, coding) induce severe expert overload in specific layers, directly reducing system throughput. Identifying and mitigating these imbalances is essential for scalable, efficient MoE inference.
\end{tcolorbox}

\subsubsection{Workload-aware Activation}
\begin{figure}[!t]
    \centering
    \includegraphics[width=1\linewidth]{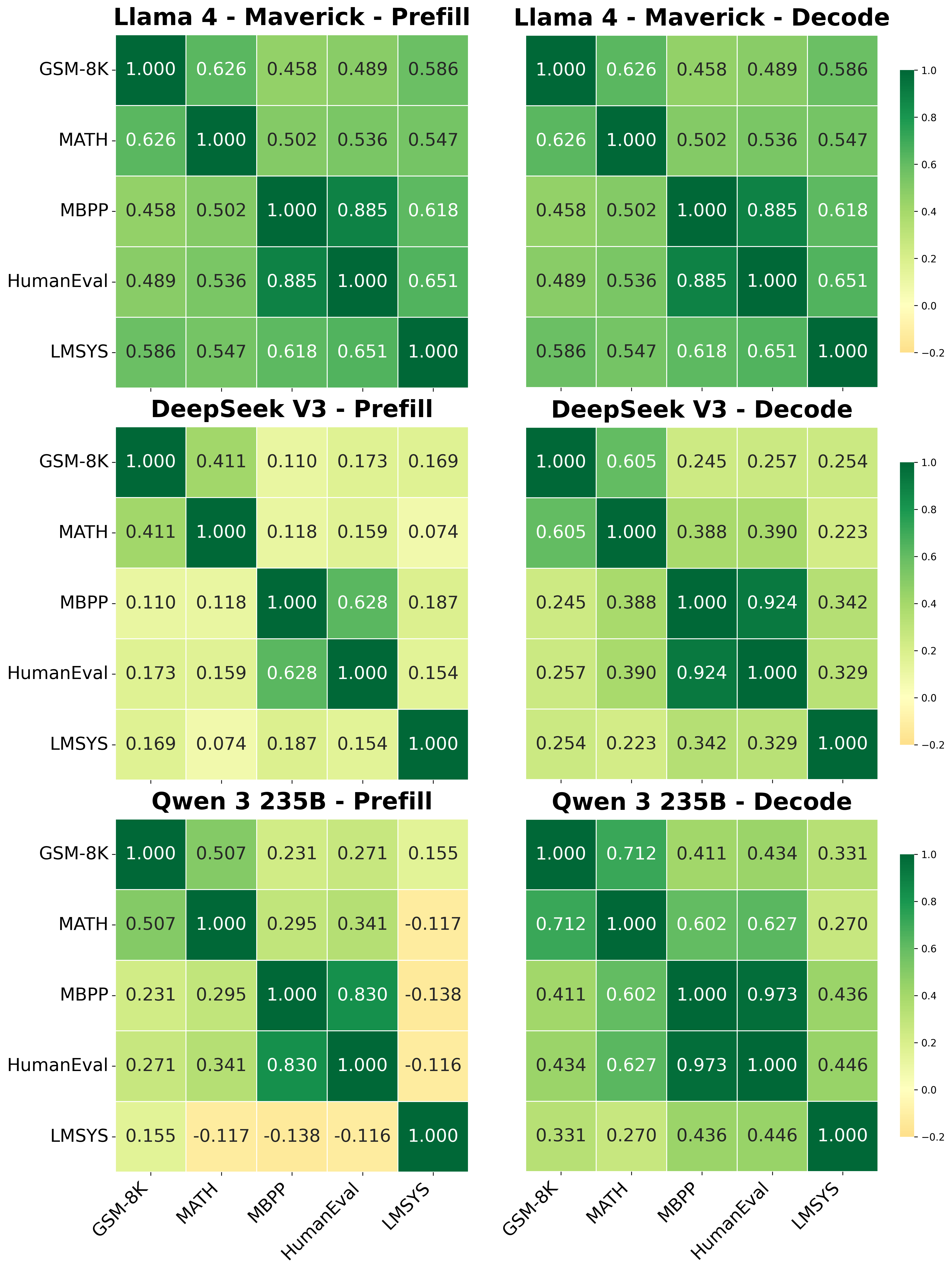}
    \caption{Correlation of expert activation patterns across datasets for Llama 4-Maverick, DeepSeek V3, and Qwen 3 235B during prefill and decode phases. Each heatmap shows the Pearson correlation between per-expert token counts for different dataset pairs.}
    \label{fig:dataset_correlation}
\end{figure}

We analyze the correlation of expert activation patterns across datasets, using Pearson correlation over per-expert token counts. As shown in \autoref{fig:dataset_correlation}, datasets within similar domains (e.g., MATH and GSM8K; HumanEval and MBPP) exhibit high correlation in the experts they activate, especially in decode stage. Notably, Maverick also shows substantial cross-domain correlation (e.g., MATH and MBPP). This might indicate that certain experts are general and are frequently activated by diverse workloads.
In contrast, DeepSeek V3 demonstrates strong within-domain correlation but low cross-domain correlation, suggesting \textit{more effective expert specialization}. For Qwen 3 235B, we found a subset of experts are highly activated across datasets, \textit{likely due to its top-8 routing mechanism without shared expert, which results in a few ``hot" experts acting as de facto shared experts}.
These patterns highlight that both model architecture and dataset characteristics drive expert activation overlap, with implications for specialization, efficiency, and potential interference between tasks.


\begin{tcolorbox}[colback=green!10!white, colframe=green!80!black, title=Insight: Workload-dependent Expert Correlation]
\textbf{Expert activation correlation is strongly workload-dependent.} In the decode stage, intra-domain correlation is high, enabling predictable batching. DeepSeek V3 achieves better domain separation via expert specialization, while Qwen 3 235B's routing leads to a few universally hot experts. Model architecture and data both drive these effects.
\end{tcolorbox}

\subsubsection{Prefill to Decode Expert Activation Correlation}
\begin{figure}[!bthp]
    \centering
    \includegraphics[width=1\linewidth]{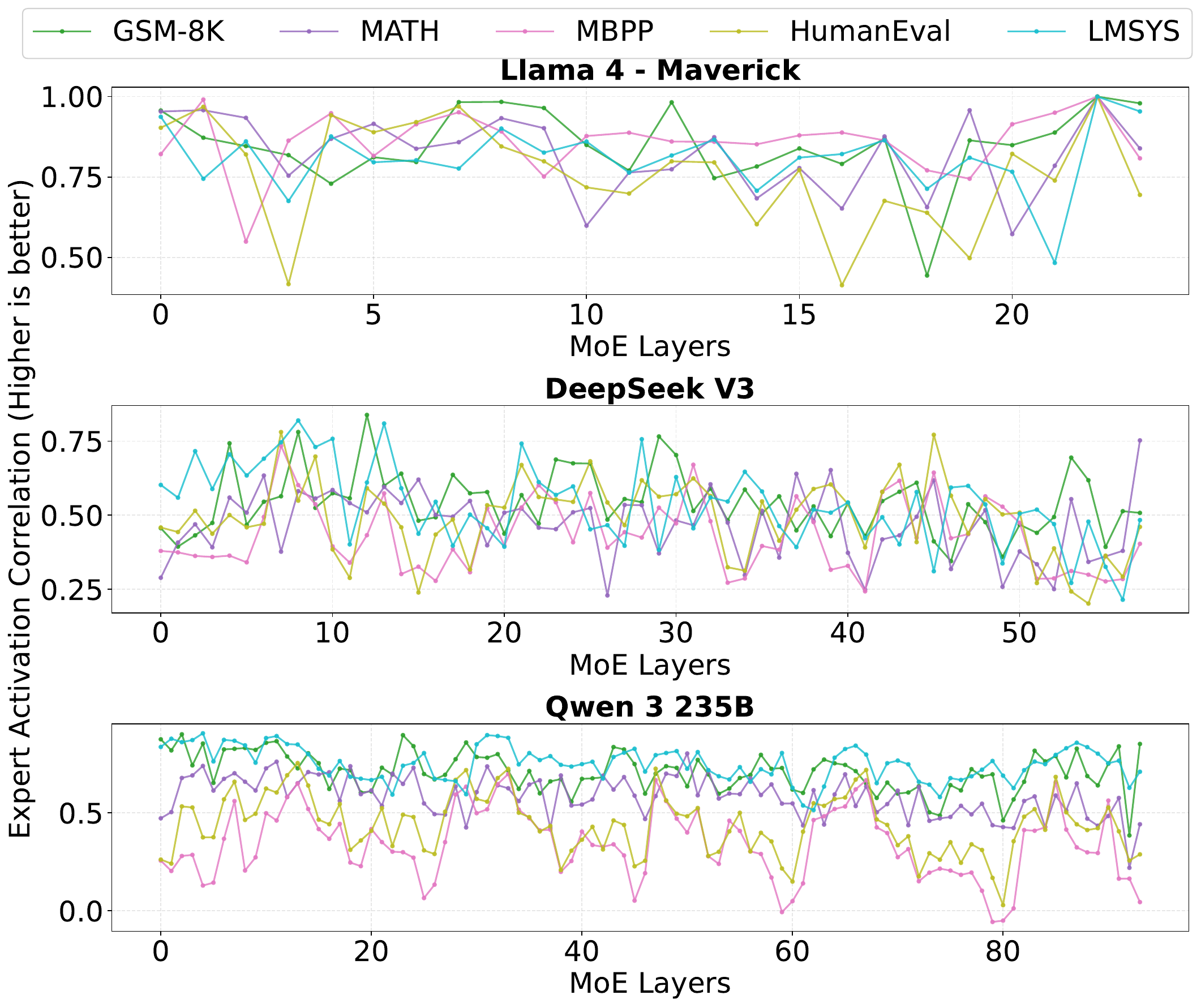}
    \caption{Expert Activation Correlation between Prefill and Decode stage  across MoE layers for different models  and datasets.}
    \label{fig:PD_correlation}
\end{figure}

To assess the predictability of expert activation between inference stages, we compute the Pearson correlation between per-expert activation vectors for prefill and decode, averaged across all experts for each layer. As shown in \autoref{fig:PD_correlation}, most layers exhibit substantial correlation, indicating that the set of experts activated during prefill is often similar to those activated during decode.
Quantitatively, the average layerwise correlation is high for Llama 4-Maverick (0.82), moderate for Qwen 3 235B (0.68), and DeepSeek V3 (0.55). While not all layers show strong correlation, the overall trend suggests that expert activation is reasonably predictable across stages, especially in models with more stable routing.
This predictability has direct implications for system optimization. In scenarios where expert weights are offloaded to remote memory or fewer GPUs are used, high prefill–decode correlation enables efficient expert prefetching and reduces latency due to expert loading. Layers with high correlation can be prioritized for prefetch reducing prefetching penalty~\citep{eliseev2023fastinferencemixtureofexpertslanguage,bambhaniya2024moeeras,xue2024moeinfinity}.
\begin{tcolorbox}[colback=green!10!white, colframe=green!80!black, title=Insight: Prefill–Decode Predictability]
\textbf{Expert activation patterns between prefill and decode are strongly correlated} in most layers, especially in Maverick. This predictability enables more efficient system-level optimizations, such as expert prefetching and memory management, and should be leveraged in large-scale MoE inference deployments.
\end{tcolorbox}

\subsubsection{Clustering Expert Activations}
\begin{figure}[!t]
    \centering
    \subfigure[Prefill Clustering.]{\includegraphics[width=0.45\columnwidth]{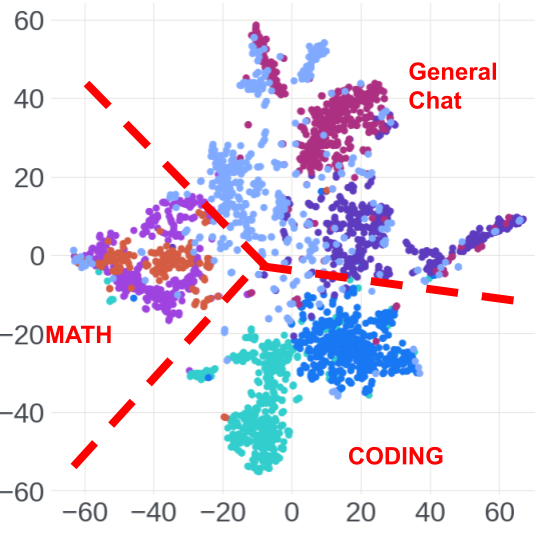}} 
    \subfigure[Decode Clustering.]{\includegraphics[width=0.48\columnwidth]{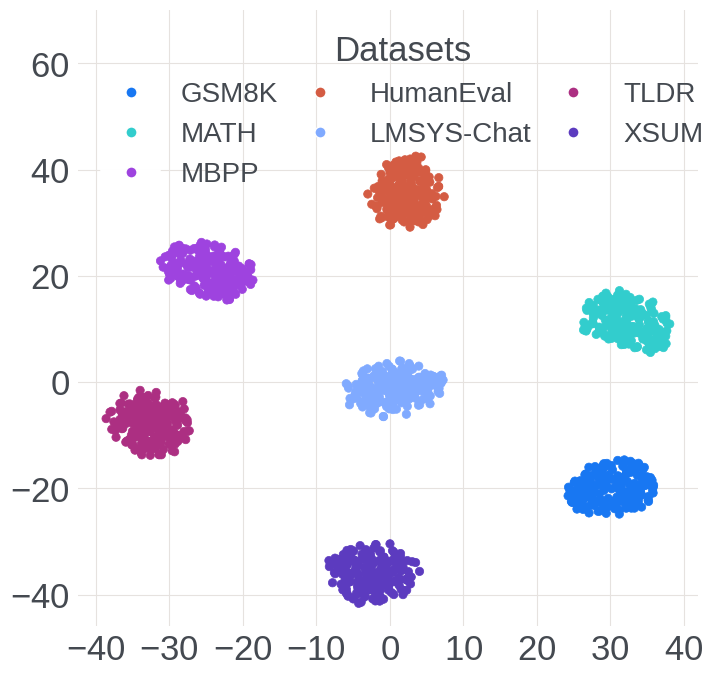}} 
    \caption{Llama-Maverick(Layer 17)'s expert activation  t-SNE visualization for request of different dataset.}
    \label{fig:maverick_clustering}
\end{figure}

To analyze request-level expert activation patterns, we apply t-SNE\footnote{t-distributed Stochastic Neighbor Embedding reduces high-dimensional data to two or three dimensions by minimizing the divergence between probability distributions of pairwise similarities in the original and embedded spaces; the axes in a t-SNE plot do not have intrinsic meaning and are used only for visualization.} to the per-request expert activation vectors. For each request, we record the number of tokens routed to each expert during both prefill and decode stages, resulting in a data matrix of shape [R, E] (requests $\times$ experts). T-SNE is then used to project these high-dimensional activation patterns into two dimensions for visualization.

In \autoref{fig:maverick_clustering}, we observe clustering of requests from similar domains during the prefill stage: requests belonging to the same dataset exhibit similar expert activation profiles, not just at the token level but across the entire prompt. This effect is even more pronounced in the decode stage, where requests from the same dataset are tightly grouped, indicating highly consistent expert activation patterns.

These findings have two major implications. First, batching requests from similar datasets can lead to predictable and concentrated expert activation, which is advantageous for system acceleration and resource allocation. Second, and more critically, the clustering reveals a potential security vulnerability: even without user-identifying tokens, it is possible to infer the request type solely from expert activation patterns. This opens the door to adversaries monitoring expert activations to request profiling.

\begin{tcolorbox}[colback=green!10!white, colframe=green!80!black, title=Insight: Predictable Clustering]
\textbf{Expert activation clustering enables robust request based grouping.}
Requests from similar domains form clear clusters in expert activation space, especially during decode. Thus, prompts can be grouped and batched efficiently by analyzing activation vectors alone, improving resource utilization and throughput.
\end{tcolorbox}
\begin{tcolorbox}[colback=red!10!white, colframe=red!80!black, title=Security Risk: Activation-based Request Profiling]
\textbf{Expert activation patterns can leak request type and user intent.}
The same clustering that aids batching also creates a security risk: activation vectors seems highly indicative of request type. Adversaries with access to this data could infer prompt categories or user intent, even without explicit identifiers, raising risks of user identity and information leakage.
\end{tcolorbox}

\section{Workload-aware Micro-Batch Grouping and Expert Placement Strategy}
\label{sec:optimizations}

In distributed Mixture-of-Experts (MoE) systems, efficiently partitioning experts across devices while maintaining high performance requires careful consideration of both request patterns and expert utilization. We propose a two-stage approach: (1) clustering requests based on expert activation patterns, and (2) strategically placing experts across device groups with controlled redundancy to optimize coverage and load balancing.

\subsection{Workload-aware Micro-batch Grouping}
\label{subsec:microbatch_grouping}
\subsubsection{Clustering Methodology}

Using past dataset with $N$ requests, where each request $r_i$ has an associated expert activation vector $\mathbf{a}_i \in \mathbb{R}^{E}$ representing the tokens sent to each of $E$ experts during decoding, we perform the following clustering procedure:

\begin{enumerate}
    \item \textbf{Normalization}: We apply L2 normalization to each activation vector to focus on relative activation patterns:
    \begin{equation}
        \tilde{\mathbf{a}}_i = \frac{\mathbf{a}_i}{\|\mathbf{a}_i\|_2}
    \end{equation}
    
    \item \textbf{K-Means Clustering}: We apply K-Means clustering to partition requests into $K$ clusters based on their normalized activation patterns:
    \begin{equation}
        \min_{\mathcal{C}} \sum_{k=1}^{K} \sum_{i \in C_k} \|\tilde{\mathbf{a}}_i - \boldsymbol{\mu}_k\|_2^2
    \end{equation}
    where $\mathcal{C} = \{C_1, \ldots, C_K\}$ represents the cluster assignments and $\boldsymbol{\mu}_k$ is the centroid of cluster $C_k$.
\end{enumerate}

This clustering approach groups together requests with similar expert usage patterns, enabling more efficient expert placement decisions which helps in reducing all-to-all communication data resulting in faster decodes.

\subsubsection{Cluster-to-Group Assignment}

For $DP$ number of expert groups, we assign the $K$ request clusters to $D$ expert groups. Our approach handles three cases:

\paragraph{Case 1: One-to-One Mapping ($K = D$)} 
Each request cluster is assigned to exactly one expert group:
\begin{equation}
    \text{map}(C_k) = \{G_k\}, \quad k \in [1, K]
\end{equation}

\paragraph{Case 2: Multiple Groups per Cluster ($D > K$)}
When we have more expert groups than request clusters, larger clusters receive multiple groups to provide better parallelism for high-traffic patterns. We:
\begin{enumerate}
    \item Compute cluster sizes: $s_k = \sum_{i \in C_k} \|\mathbf{a}_i\|_1$
    \item Sort clusters by size: $C_{\pi(1)}, \ldots, C_{\pi(K)}$ where $s_{\pi(1)} \geq \cdots \geq s_{\pi(K)}$
    \item Assign one group to each cluster, then distribute the remaining $D - K$ groups to the largest clusters in round-robin fashion
\end{enumerate}

\paragraph{Case 3: Multiple Clusters per Group ($K > D$)}
When we have more request clusters than expert groups, we perform secondary clustering to merge request clusters into $D$ meta-clusters:
\begin{equation}
    \mathbf{u}_k = \sum_{i \in C_k} \mathbf{a}_i, \quad k \in [1, K]
\end{equation}
We then apply K-Means with $D$ clusters on $\{\mathbf{u}_1, \ldots, \mathbf{u}_K\}$ to assign multiple request clusters to each expert group.

\subsection{Data-based Expert Placement}
\label{subsec:expert_placement}

\subsubsection{Problem Formulation}

Given $E$ unique experts and $D$ expert groups, with a redundancy budget of $R$ experts, we must place experts such that:
\begin{itemize}
    \item Each group contains exactly $M = \frac{E + R}{D}$ experts
    \item All $E$ unique experts are covered across all groups
    \item Expert placement maximizes relevance to assigned request clusters
\end{itemize}

For each expert group $G_d$, we compute the aggregated expert usage based on its assigned clusters:
\begin{equation}
    U_{d,e} = \sum_{k: d \in \text{map}(C_k)} \sum_{i \in C_k} a_{i,e}
\end{equation}
where $a_{i,e}$ is the activation count of expert $e$ for request $i$.

\subsubsection{Two-Phase Expert Placement Algorithm}

\paragraph{Phase 1: Unique Expert Distribution}

We first distribute all $E$ unique experts across the $D$ groups while balancing coverage and relevance. Let $M_{\min} = \lfloor E/D \rfloor$ and $M_{\max} = \lceil E/D \rceil$ denote the minimum and maximum unique experts per group.

\begin{algorithm}[h]
\caption{Phase 1: Unique Expert Distribution}
\begin{algorithmic}[1]
\STATE Compute global expert importance: $I_e = \max_{d \in [1,D]} U_{d,e}$
\STATE Sort experts by importance: $e_{\pi(1)}, \ldots, e_{\pi(E)}$ where $I_{\pi(1)} \geq \cdots \geq I_{\pi(E)}$
\FOR{each expert $e$ in sorted order}
    \STATE Find groups sorted by usage: $d_1, \ldots, d_D$ where $U_{d_1, e} \geq \cdots \geq U_{d_D, e}$
    \FOR{each group $d$ in sorted order}
        \IF{$|G_d| < M_{\max}$}
            \STATE $G_d \leftarrow G_d \cup \{e\}$
            \STATE \textbf{break}
        \ENDIF
    \ENDFOR
\ENDFOR
\end{algorithmic}
\end{algorithm}

This greedy approach assigns each expert to the group that uses it most, subject to capacity constraints, ensuring balanced distribution.

\paragraph{Phase 2: Redundant Expert Addition}

After distributing unique experts, we fill remaining slots in each group with redundant copies of experts to reach the target size $M$:

\begin{algorithm}[h]
\caption{Phase 2: Redundant Expert Addition}
\begin{algorithmic}[1]
\FOR{each group $d \in [1, D]$}
    \STATE $\text{needed} \leftarrow M - |G_d|$
    \STATE $\text{candidates} \leftarrow [E] \setminus G_d$ \COMMENT{Experts not in group}
    \STATE Sort candidates by usage: $c_1, \ldots, c_{|\text{candidates}|}$ where $U_{d,c_1} \geq \cdots \geq U_{d,c_{|\text{candidates}|}}$
    \STATE $G_d \leftarrow G_d \cup \{c_1, \ldots, c_{\text{needed}}\}$
\ENDFOR
\end{algorithmic}
\end{algorithm}

This phase adds the most relevant experts that weren't initially assigned to each group, creating controlled redundancy that improves cache hit rates for frequently co-activated experts. While we enable redundant experts, for the evaluation we don't use any redundant experts.

\subsubsection{Group Balancing and Verification}

After the two-phase placement, we perform balancing to ensure exact size constraints:

\begin{itemize}
    \item \textbf{Undersized groups} ($|G_d| < M$): Randomly add missing experts from those not yet in $G_d$
    \item \textbf{Oversized groups} ($|G_d| > M$): Remove redundant experts (those appearing in multiple groups) from the end of the sorted list, preserving unique experts
\end{itemize}

We verify two critical properties:
\begin{equation}
    \bigcup_{d=1}^{D} G_d = [E], \quad |G_d| = M \quad \forall d \in [1, D]
\end{equation}

The first ensures complete coverage of all experts, while the second guarantees balanced memory usage across devices.

\subsubsection{Request Routing}

Given the expert placement, each request $r$ is routed to its recommended expert groups based on its cluster assignment:
\begin{equation}
    \text{route}(r) = \text{map}(C_k) \quad \text{where } r \in C_k
\end{equation}

This ensures that requests are processed by groups containing the experts they are most likely to activate, minimizing cross-node all-to-all communication.

\subsection{Deployment in Production Stack}
\begin{figure}[!bthp]
    \centering
    \includegraphics[width=1\linewidth]{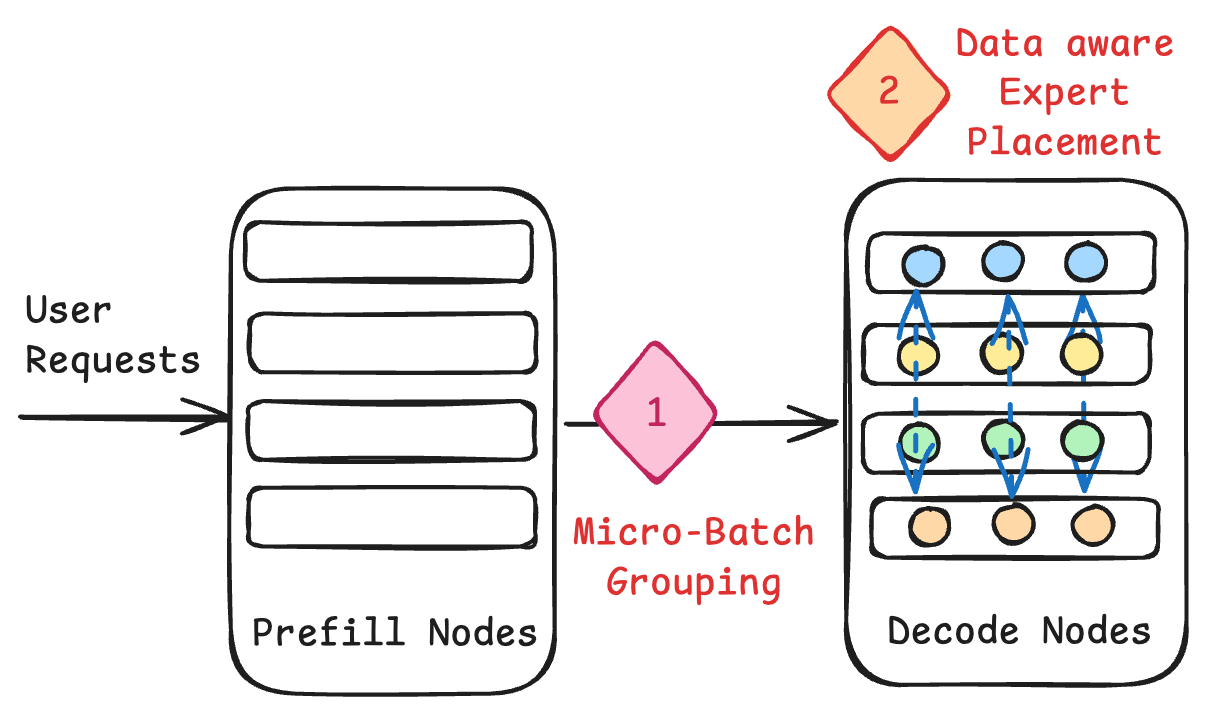}
    \caption{Disaggregated Prefill-Decode production system architecture with workload-aware micro-batch grouping and expert placement strategy}
    \label{fig:disagg_flow}
\end{figure}

To implement the proposed clustering and placement strategy, in production. The MoE inference stack will need to be modified as shown in \autoref{fig:disagg_flow}.

\circled{1} \textbf{Micro-Batch Grouping}: Analyzing which expert are being activated by the tokens of the request. Using this information we decide the dataset of the request and which decode micro batch it should be placed in. This will happen each time a request comes in.

\circled{2} \textbf{Data-aware Expert Placement}: Decode expert placement is dictated by finding experts with correlated activation patterns. We place these experts in the same node, to minimize inter-node all-to-all communication. New expert placement should happen infrequently, the main reason for refreshing expert placement would be intra-node expert corelation has droped resulting in additional inter-node all-to-all communication.

\section{Evaluations}
\label{sec:evaluation}

In this section, we rigorously evaluate the impact of workload-aware micro-batch grouping and expert placement strategies on accelerating multi-node MoE inference. Our evaluation focuses on three key aspects: (1) the ability to accurately classify prefill request type for micro-batch formation, (2) reduction in all-to-all communication volume via data-aware expert placement, and (3) improvements in MoE layer latency.

 We demonstrate the efficacy of our approach using three state-of-the-art MoE models: Llama-Maverick, DeepSeek-V3, and Qwen 3-235B, evaluated on the same diverse datasets as in ~\autoref{sec:characterization}.

\subsection{Prefill Request Type Classification}
\label{sec:prefill-classification}

In \autoref{fig:maverick_clustering}, we saw that prefill requests show significant clustering of request based on expert activation. To quantify this clustering of request types based on expert activation, we perform dataset classification for each MoE layer independently. For each model, we construct a request-wise expert activation matrix of shape $[R, E]$, where $R$ is the total number of requests and $E$ is the number of experts. Each entry records the raw token count routed to each expert for a given request. The dataset is split into 80:20 train:test sets, with equal representation of each dataset.
We train a multi-class logistic regression classifier for each layer, using the raw expert activation counts as input features.

\begin{figure*}[!bthp]
    \centering
    \includegraphics[width=0.8\linewidth]{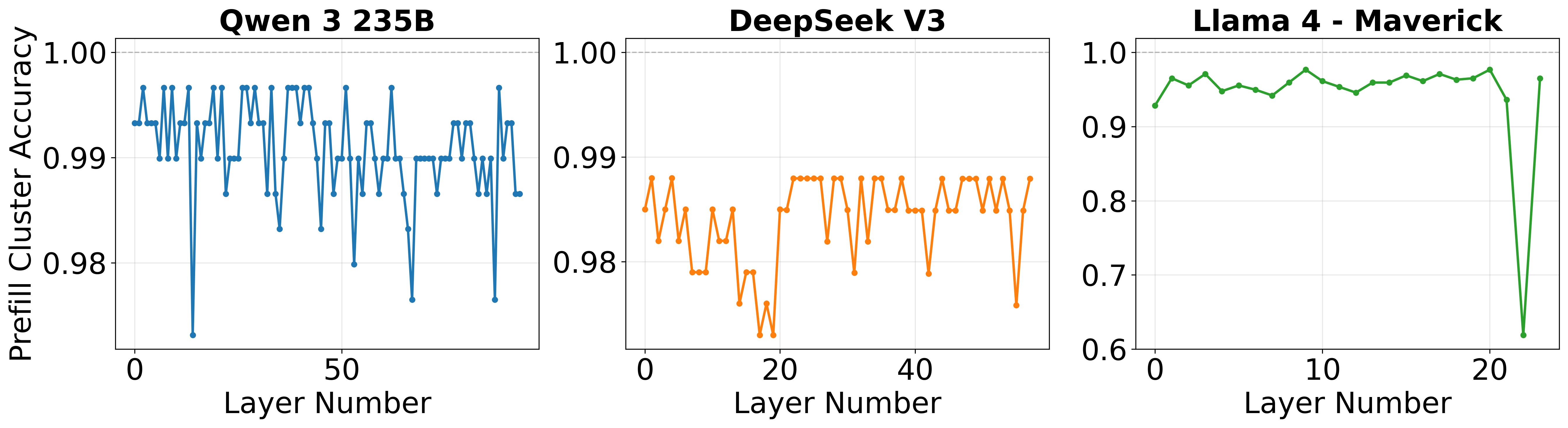}
    \caption{Accuracy for classifying prefill request into clusters based on the expert activation patterns.}
    \label{fig:prefill_cluster_accuracy}
\end{figure*}

\autoref{fig:prefill_cluster_accuracy} presents the layer-wise classification accuracy for all three models. The average classification accuracy is exceptionally high: Qwen3-235B achieves $98.9\% \pm 0.4$, DeepSeek-V3 achieves $98.4\% \pm 0.4$, and Llama-Maverick achieves $94.4\% \pm 6.8$ across layers. These results demonstrate that prefill expert activation patterns are highly predictive of request type, enabling robust micro-batch formation for subsequent decode stages.

\subsection{Inter-Node All-to-All Communication}
\label{sec:all2all}
\begin{figure}[!bthp]
    \centering
    \includegraphics[width=1\linewidth]{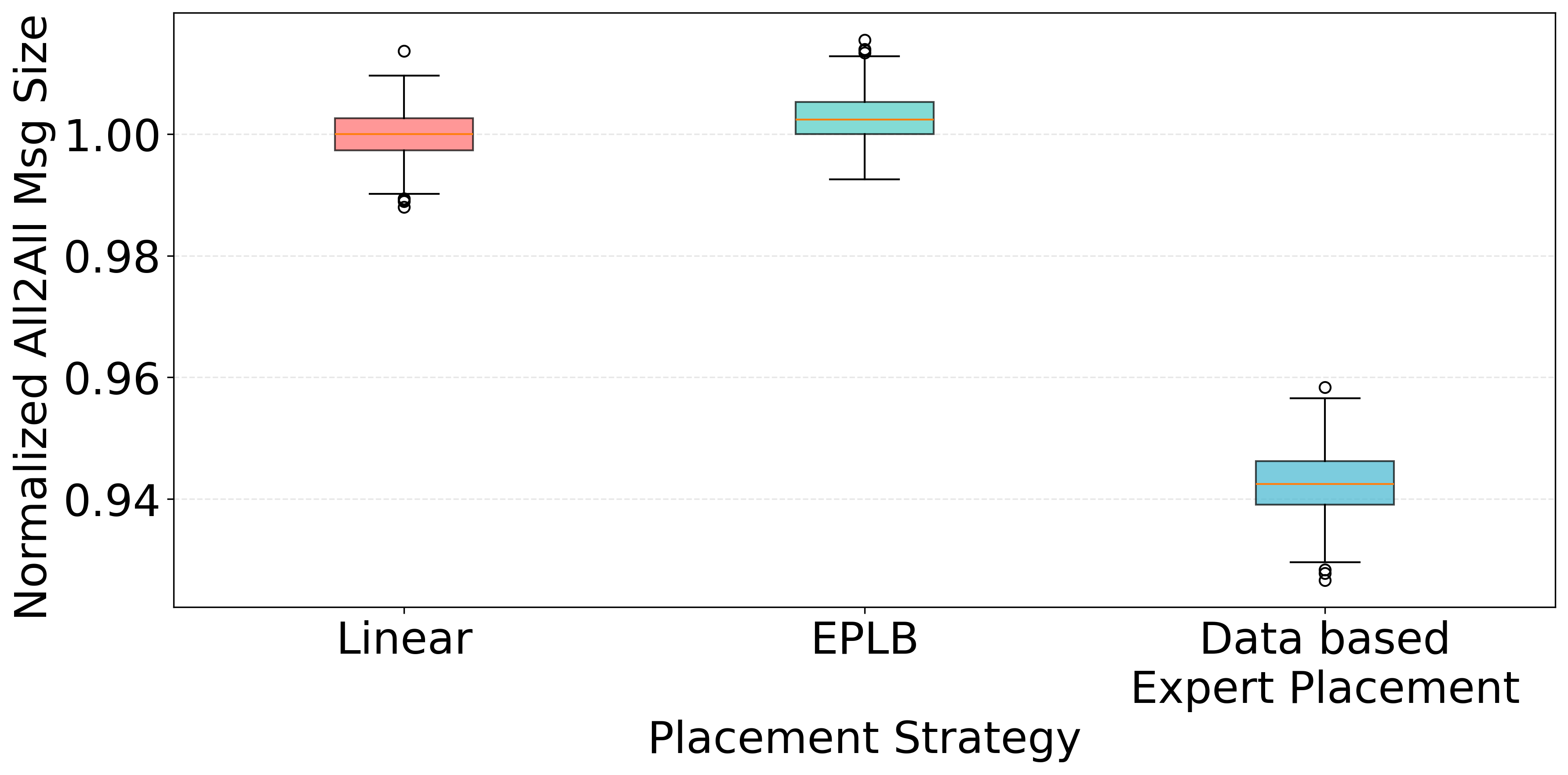}
    \caption{Normalized all-to-all data size for Deepseek-V3 (Layer 42) with different expert placement strategies.}
    \label{fig:deepseek_all2all_data}
\end{figure}

We next evaluate the impact of our proposed workload-aware micro-batch grouping and expert placement on inter-node all-to-all communication volume during MoE Decode. Accurate prefill request type prediction enables micro-batch formation, and our data-driven expert placement strategy (\textit{Data-based Expert Placement}) co-locates frequently co-activated experts in the same node, reducing the need for tokens to be routed across nodes.
We compare three expert placement strategies on DeepSeek-V3 with a DP8TP8$\rightarrow$EP64 configuration, averaged over 500 global batches (each global batch consists of $8 \times 128$ requests):
\begin{itemize}
    \item \textbf{Linear}~\citep{vLLM_expert_placement_2025}: Experts are assigned contiguously to ranks (e.g., rank 0: [0,1], rank 1: [2,3]).
    \item \textbf{Expert Placement with Load Balancing (EPLB)}~\citep{deepseek_EPLB_2025_github, efficientMoE_inference_fair}: Expert placement based on past expert loads.\footnote{EPLB also has redundant experts, but for our analysis we have 0 redundant experts for all placement experts.}
    \item \textbf{Data-based Expert Placement}: Our proposed method, which clusters requests and places correlated experts on the same node.
\end{itemize}
All-to-all data volume is normalized to the median value of the linear baseline. \autoref{fig:deepseek_all2all_data} shows the normalized all-to-all data size for each strategy. Data-based Expert Placement achieves a substantial reduction in inter-node communication, with a median normalized value of $0.94$, compared to $1.00$ for Linear and $1.00$ for EPLB. This reduction is due the fact the more tokens are located in the same node as the destination experts, thus reduce inter-node communication data.

In summary, data-aware expert placement significantly reduces all-to-all communication volume, directly addressing a key bottleneck in scaling multi-node MoE inference.

\subsection{MoE Decode Runtime}
\label{sec:moe-runtime}
Finally, we evaluate the impact of Data-based Expert Placement combined with workload-aware micro-batch grouping on MoE layer runtime. We present results for two representative models: Llama Maverick and Qwen3-235B. Both using a DP8TP8$\rightarrow$EP64 configuration, with each global batch consisting of $8 \times 128$ requests.

\begin{figure}[!bthp]
    \centering
    \includegraphics[width=1\linewidth]{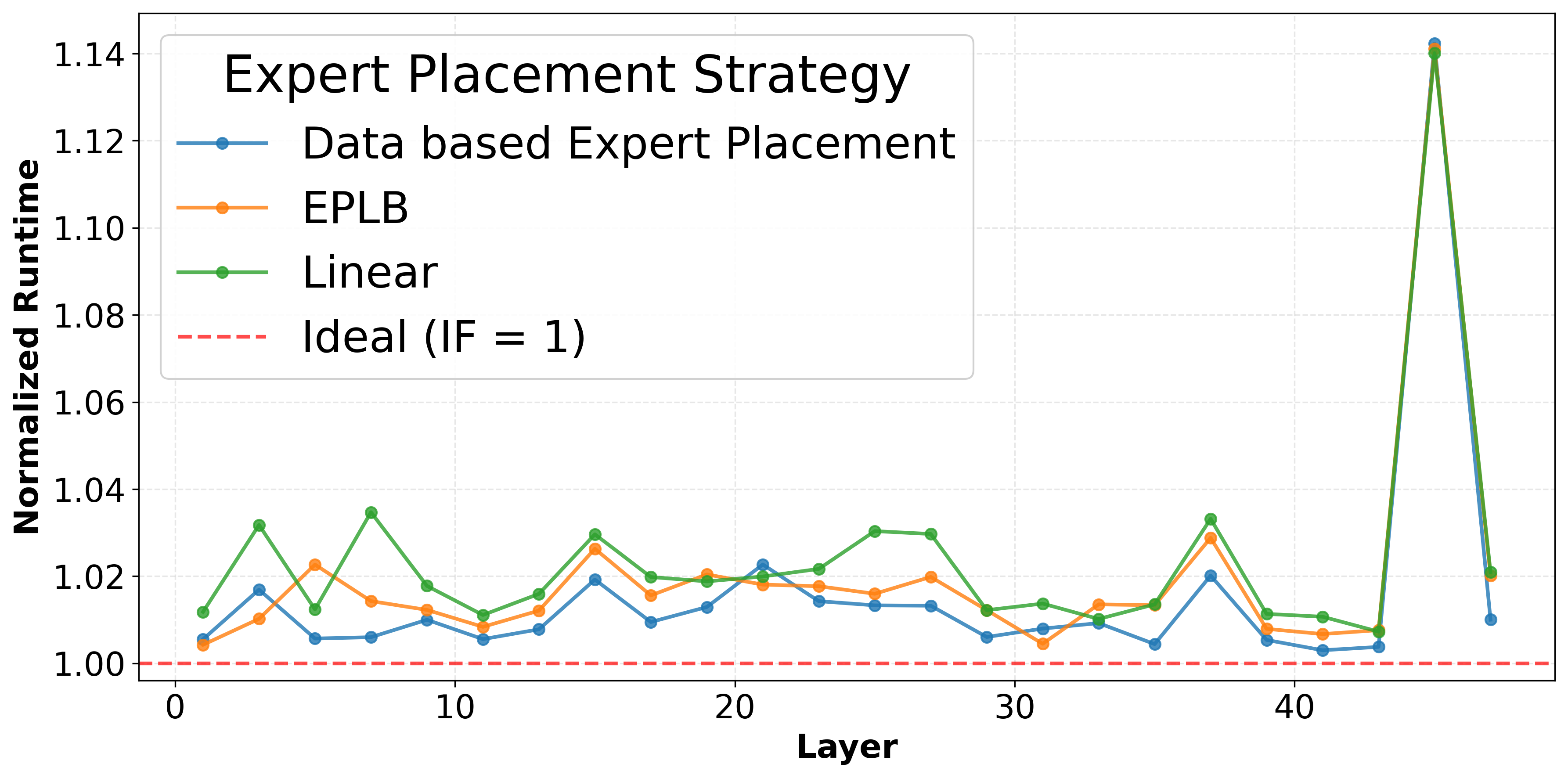}
    \caption{Runtime for MoE Layer of Llama-Maverick with different expert placement strategies.}
    \label{fig:llama_layerwise_runtime}
\end{figure}

For Llama Maverick, \autoref{fig:llama_layerwise_runtime} shows the median layer runtime over 100 global batches for different expert placement strategies. Data-based Expert Placement outperforms EPLB (Expert Placement with Load Balancing) in most layers, achieving up to a 20\% reduction in all-to-all message size. However, due to the inefficiency of the underlying all-to-all communication kernel\footnote{Current all-to-all kernel implementation pads the data from each EP rank to the largest EP rank all-to-all data size and thus does not fully exploit reduced all-to-all message size, limiting runtime improvements.}, the observed reduction in all-to-all communication time is 9\%, which translates to a 5.5\% reduction in overall layer runtime.

\begin{figure}[!bthp]
    \centering
    \includegraphics[width=1\linewidth]{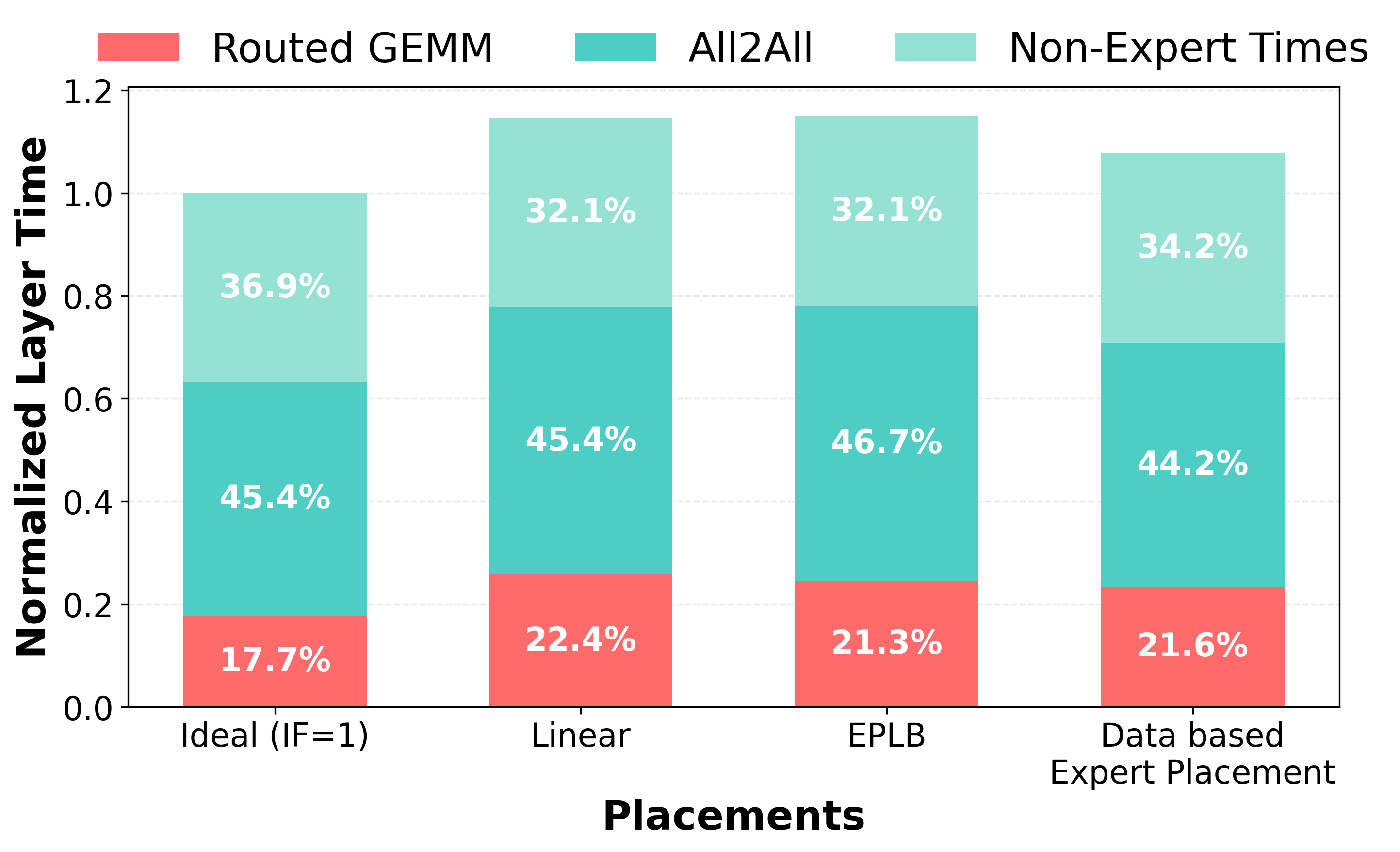}
    \caption{Runtime Breakdown for Qwen3-235B-A22B with different expert placement strategies.}
    \label{fig:qwen_runtime_breakdown}
\end{figure}

We further analyze Qwen3-235B, focusing on a single representative layer. \autoref{fig:qwen_runtime_breakdown} presents the runtime breakdown of the median layer runtime over 100 global batches for different placement strategies. Data-based Expert Placement again demonstrates clear benefits, reducing all-to-all communication time by 12\% and overall layer runtime by 6\% relative to EPLB.

These results reaffirm that by routing more tokens to local experts and minimizing inter-node communication, Data-based Expert Placement can delivers consistent improvements in MoE layer efficiency. The magnitude of runtime reduction is ultimately bounded by the efficiency of the underlying communication primitives, highlighting an important direction for future system optimization.

\section{Related Work}
\label{sec:related}
Recent work on Mixture-of-Experts (MoE) inference has explored several optimization strategies. Huang et al.~\citep{huang2024toward} focus on routing optimizations, token droping and suggests Greedy placement, while Lu et al.~\citep{lu-etal-2024-experts} propose expert pruning and skipping to improve efficiency. 
Deepspeed-MoE~\citep{rajbhandari2022deepspeedmoeadvancingmixtureofexpertsinference} introduced hybrid parallelism, enabling large multi-node MoE deployments. Expert Affinity~\citep{yao2024exploitinginterlayerexpertaffinity} reduce cross-GPU communication via KV cache duplication, but this approach is not scalable as cache sizes grow. MoETuner~\citep{go2025moetuneroptimizedmixtureexpert} uses integer linear programming for expert placement, but is limited to small expert counts.
MoESys~\citep{MoESYS2024} proposes an Elastic MoE training strategy with 2D prefetch and Fusion communication over Hierarchical storage, so as to enjoy efficient parallelisms.

DeepSeek-V3~\citep{deepseekai2025deepseekv3technicalreport} introduced EPLB, periodically rebalancing and replicating experts to improve throughput. Deepseek-V3 also proposed overlapping the attention of one micro-batch with the dispatch+MoE+combine of another to increase utilization. Our proposed placement would also work with this optimization.
Despite these advances, scalable multi-node MoE inference remains bottlenecked by communication overhead and lacks activation-aware batching and placement, which our work directly addresses.

\section{Conclusion}
\label{sec:conclusion}
This work addresses a central bottleneck in scaling multi-node Mixture-of-Experts (MoE) inference: the mismatch between dynamic expert activation patterns and static device topology, which leads to excessive inter-node communication and suboptimal runtime. Through comprehensive data-driven analysis across multiple state-of-the-art MoE models and diverse datasets, we demonstrate that expert activation patterns are highly predictive of request type and exhibit strong clustering properties.

Building on these insights, we introduce a workload-aware micro-batch grouping and data-based expert placement strategy that co-locates frequently co-activated experts and batches similar requests. Our evaluations show that this approach enables highly accurate prefill request type classification, substantially reduces all-to-all communication volume, and delivers consistent improvements in MoE layer runtime across models and layers. Specifically, we observe up to 20\% reduction in inter-node message size and up to 6\% reduction in layer runtime, with the magnitude of improvement bounded by the efficiency of underlying communication kernels.

These results establish a principled foundation for scaling sparse model inference in distributed environments. By leveraging activation-aware batching and expert placement.

\textbf{Future Works and Extensions:} Further optimization of all-to-all communication kernel where padding is not required can help reduce the all-to-all Latency. Using expert load information during expert grouping can further improve the overall layer latency taking it more closer to ideal runtime.

\clearpage

\bibliography{references}
\bibliographystyle{mlsys2025}

\clearpage

\end{document}